\documentclass[lettersize,journal]{IEEEtran}
\usepackage{amsmath,amsfonts}
\usepackage{algorithmic}
\usepackage{algorithm}
\usepackage{array}
\usepackage[caption=false,font=footnotesize,labelfont=sf,textfont=sf]{subfig}
\usepackage{textcomp}
\usepackage{stfloats}
\usepackage{url}
\usepackage{hyperref}
\usepackage{verbatim}
\usepackage{graphicx}
\usepackage{float}
\usepackage{subfig}
\usepackage{caption}
\usepackage{booktabs}
\usepackage{array}
\usepackage{xcolor}
\usepackage{multirow}
\usepackage{subcaption}
\usepackage{cite}
\usepackage{marvosym}

\begin{document}

\title{Continual All-in-One Adverse Weather Removal with Knowledge Replay on a Unified Network Structure}

\author{
    De~Cheng,
    Yanling~Ji,
    Dong~Gong,
    Yan Li,
    Nannan~Wang,~\IEEEmembership{Senior Member,~IEEE,}
    Junwei~Han,~\IEEEmembership{Fellow,~IEEE,}
    Dingwen~Zhang.
    

\thanks{D.~Cheng, N.~Wang are with the School of Telecommunications Engineering, Xidian University, Xi'an 710071, Shaanxi, P. R. China (email: dcheng@xidian.edu.cn, and nnwang@xidian.edu.cn)}
\thanks{Y.~Ji, J.~Han are with the School of Automation, Northwestern Polytechnical University, Xi'an, Shaanxi, P. R. China (email: jiyanling6@gmail.com, junweihan2010@gmail.com.)}
\thanks{D.~Zhang is with the School of Automation, Northwestern Polytechnical University, Xi'an, Shaanxi, P. R. China, and Hefei Comprehensive National Science Center, Hefei, Anhui, 230088, China (email: zhangdingwen2006yyy@gmail.com.)}
\thanks{D.~Gong is with the University of New South Wales (UNSW), Sydney, Australia. (e-mail: edgong01@gmail.com).} 
\thanks{Y.~Li is with Shandong Normal University, Jinan, China. (e-mail: yanli.ly.cs@gmail.com).} 
\thanks{De~Cheng and Yanling Ji contribute equally to this work.} 
\thanks{corresponding author: Dingwen Zhang.}
}

\markboth{IEEE Transactions on Multimedia}%
{Shell \MakeLowercase{\textit{et al.}}: A Sample Article Using IEEEtran.cls for IEEE Journals}


\maketitle

\begin{abstract}

In real-world applications, image degeneration caused by adverse weather is always complex and changes with different weather conditions from days and seasons. Systems in real-world environments constantly encounter adverse weather conditions that are not previously observed. Therefore, it practically requires adverse weather removal models to continually learn from incrementally collected data reflecting various degeneration types. Existing adverse weather removal approaches, for either single or multiple adverse weathers, are mainly designed for a static learning paradigm, which assumes that the data of all types of degenerations to handle can be finely collected at one time before a single-phase learning process. They thus cannot directly handle the incremental learning requirements. To address this issue, we made the earliest effort to investigate the continual all-in-one adverse weather removal task, in a setting closer to real-world applications. Specifically, we develop a novel continual learning framework with effective knowledge replay (KR) on a unified network structure. Equipped with a principal component projection and an effective knowledge distillation mechanism, the proposed KR techniques are tailored for the all-in-one weather removal task. It considers the characteristics of the image restoration task with multiple degenerations in continual learning, and the knowledge for different degenerations can be shared and accumulated in the unified network structure. Extensive experimental results demonstrate the effectiveness of the proposed method to deal with this challenging task, which performs competitively to existing dedicated or joint training image restoration methods. Our code is available at \href{https://github.com/xiaojihh/CL_all-in-one}{https://github.com/xiaojihh/CL\_all-in-one}.
\end{abstract}


\section{Introduction}

\begin{figure}[tbp] 
    \centering
    \includegraphics[width=\linewidth]{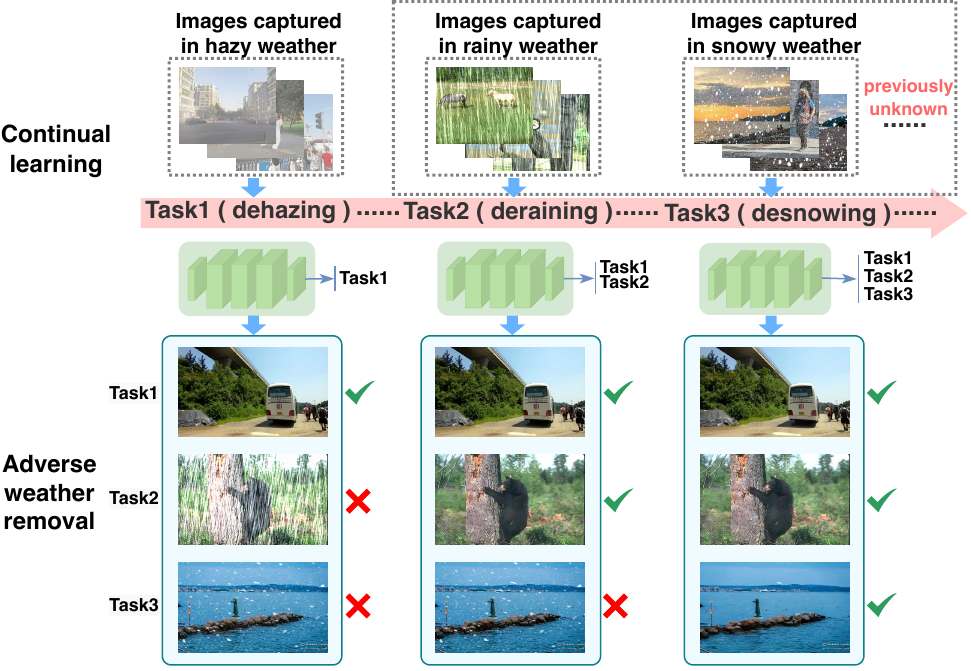}
    \caption{Illustration of the proposed continual learning for all-in-one adverse weather removal task. Left: Model trained on hazy images can only conduct dehazing; Middle: When the model is trained continually on rainy images without accessing previous data, it can well conduct dehazing and deraining simultaneously; Right: As continual learning goes on, the model can accumulate knowledge towards an all-in-one model, to deal with all types of adverse weather.  
    } 
    \label{fig::overview}
\end{figure}


\IEEEPARstart{V}{arious} types of adverse weather, for example, haze, rain, and snow, are ubiquitous in photography, which degenerate the imaging quality and thus hinder the other downstream computer vision tasks, \emph{e.g.}, object detection~\cite{od_2023}, image classification~\cite{ic_2017}, semantic segmentation~\cite{seg_2017}, and many others~\cite{more01_2020, more02_2020}. To address these issues, many single adverse weather removal tasks have been extensively studied in the past decades, including image dehazing~\cite{FFA2020, Haze01_2022}, deraining~\cite{Rain01_2019, Rain02_2022, Rain03_2021}, and desnowing~\cite{Snow01_2021, Snow02_2020}. 

 
However, most existing methods \cite{liu2021joint, Haze01_2022, Rain02_2022, shin2021region, Snow01_2021,li2019pdr, yi2021efficient} are specifically designed to handle a single type of degeneration/weather, and they can work only when the specific degeneration type is appointed. This greatly limits the deployment of such methods in real-world scenarios, such as video surveillance and autonomous vehicle systems, which require adverse weather removal methods to handle various complex weather conditions automatically. 
Although some efforts have been taken to handle multiple degenerations in an ``all-in-one'' manner  \cite{Multitask01_2022_CVPR,Multitask02_2021_CVPR,Multitask03_2020_CVPR} (i.e., using a unified deep model to deal with different types of adverse weather simultaneously), these methods are typically designed based on either of the following two assumptions: 1) All types of adverse weather removal training data need to be finely collected before the model training phase~\cite{Multitask01_2022_CVPR,Multitask02_2021_CVPR,Multitask03_2020_CVPR}; 2) In the testing phase, degeneration type is required to utilize the corresponding model component for specific adverse weather removal task~\cite{Multitask03_2020_CVPR,Multitask01_2022_CVPR}.  
Unfortunately, such assumptions prevent existing adverse weather removal models from working in real-world scenarios. Firstly, image degenerations caused by adverse weather are always complex and changing with different weather conditions from days and seasons, and the diverse manifestations of adverse weather conditions make it very challenging to collect adequate training data at one time. The systems running in real-world scenarios require the adverse weather removal method to continually learn from the incrementally collected datasets with various types of degenerations, which thus can accumulate the knowledge toward an all-in-one model. Therefore, the first assumption of static training style with all types of data available simultaneously, will pose a challenge for such all-in-one models and render them susceptible to the challenges of continual learning, resulting in limited applications in complex real-world scenarios. Besides, knowing the type of adverse weather in advance during application is always difficult, which also requires additional model components and computation costs.

Therefore, in this paper, we make an early effort to investigate the continual all-in-one adverse weather removal task towards a setting closer to real-world scenarios, as shown in Fig~\ref{fig::overview}. Such a setting is more realistic than the setting in previous relevant work PIGWM~\cite{PIGWM2021}, which performs CL only on a single type of degeneration with multiple deraining datasets. While in the newly proposed CL for an all-in-one setting, we study the CL for multiple adverse weather removal tasks including dehazing, deraining, and desnowing. To be specific, we build a benchmark setting with three large datasets under different adverse weather types, i.e., OTS~\cite{OTS2019}, Rain100H~\cite{Rain100H_2017_CVPR} and Snow100K~\cite{Snow100K2018}, to extensively validate the capability of CL approaches of our proposed method and the followings. To achieve the all-in-one process, we design a unified network structure with tailored effective knowledge replay for CL, where the unified structure with shared parameters enables the tasks to directly share knowledge and makes the model easy to use in practice (for example, not requiring degeneration type in testing). Notably, the proposed CL method can train the unified network on different types of input data, which may occur sequentially or alternatively.

\begin{figure*}[htbp] 
    \centering 
    \includegraphics[width=0.95\textwidth]{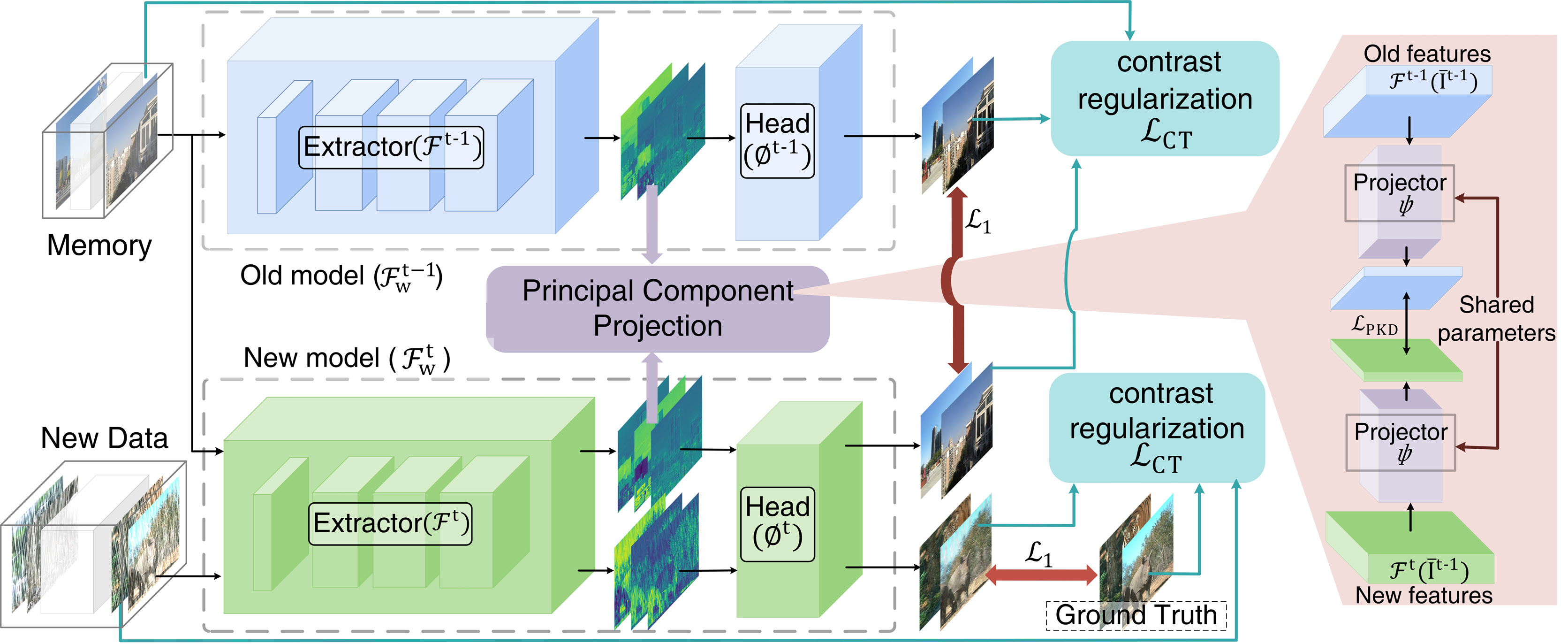} 
    \caption{The proposed continual learning framework with knowledge replay on a unified network structure for all-in-one adverse weather removal. The framework mainly includes three components: the backbone network and training scheme applied for single adverse weather removal; effective knowledge replay based distillation on the network prediction; and the principal component projection in knowledge replay.} 
    \label{fig::model}
\end{figure*}

Due to the challenging nature of the investigated continual all-in-one adverse weather removal task, we have to face a much severe catastrophic problem, since different types of adverse weather always have their own intrinsic mapping rules, and the differences among them are larger than that among different datasets while within the same type of adverse weather. To tackle such serious \emph{catastrophic forgetting} issue, we develop a unified network structure with experience replay strategy on a tiny memory buffer \cite{ER,darkexperiencereply} saving a very small subset of old samples. Unlike standard experience replay with only rehearsal on old samples in memory \cite{ER}, we design knowledge replay (KR) techniques tailored for image restoration tasks with multiple adverse weather degeneration. Considering that the all-in-one restoration task can be seen as a many-to-one feature mapping problem, the image features from different adverse weather domains (such as, hazy, rain, and snow) need to be first transformed into clean image features and then mapped into clean natural images. As clean image decoding can be shared among different types of adverse weather images, we encourage this part of knowledge among different tasks can be shared and hope the diverse representations for different degeneration can be accumulated with less interference in the mid-level representations. Thus we perform effective KR on the middle-level representations, instead of \emph{only} at the network prediction level \cite{ER,darkexperiencereply}. Specifically, we propose to perform the KR on the mid-level features in a shared latent space relying on a learned \emph{principal projection} with a lower dimension, which tends to preserve principal knowledge of different tasks to replay and accumulate, as shown in Fig. \ref{fig::model}. Moreover, we also conduct KR on the memory samples with the network's predictions (instead of their actual ground truth) as supervision, which preserve the learned knowledge in the network reflected in the network prediction. It makes training smoother and easier in practice, and enables the memory to save samples without the ground-truth.

We summarize the contributions of this paper as follows:
\begin{itemize}
\item We made the earliest effort to investigate continual learning for the all-in-one adverse weather removal task towards a setting closer to real-world scenarios. Also, we built a benchmark setting for this task, which will be open-sourced to facilitate comparison for subsequent methods on this task.

\item We develop a novel continual learning framework tailored for the all-in-one adverse weather removal task with effective knowledge replay on a unified network structure,
which replays the learned knowledge in the network with a principal component projection and an effective knowledge distillation mechanism. This helps to build a practical system to achieve multi-degradation image restoration dynamically.
\item Extensive experimental results demonstrate the effectiveness of the proposed method to deal with this challenging task, which performs competitively to existing dedicated or joint training image restoration methods. 
\end{itemize}

The rest of this paper is organized as follows. Section \ref{sec:related work} introduces related works. Section \ref{sec:proposed method} details the proposed method. Section \ref{sec:experiments} shows experimental results to demonstrate the effectiveness of the proposed method. Finally, conclusions are drawn in Section \ref{sec:conclusion}. 

\section{Related Work}\label{sec:related work}
\subsection{Adverse Weather Removal}

There exists a large amount image restoration algorithms for adverse weather removal, including image dehazing \cite{FFA2020, Haze01_2022}, deraining \cite{Rain01_2019, Rain02_2022, Rain03_2021},  desnowing \cite{Snow01_2021, Snow02_2020}, and some all-in-one methods\cite{Multitask01_2022_CVPR, Multitask02_2021_CVPR, Multitask03_2020_CVPR,LIRA}.

\textbf{Single Adverse Weather Removal.} Recent years have witnessed rapid development in the task of image restoration for adverse weather conditions based on deep learning. We will briefly introduce some representative single-weather removal methods. For dehazing, CARL~\cite{Haze01_2022} proposes to optimize the dehazing network under the consistency-regularized framework with contrast-assisted reconstruction loss, and FFA-Net \cite{FFA2020} proposes a novel feature attention module, which combines channel attention with pixel attention mechanism.
For deraining, PReNet \cite{Rain01_2019} introduces a recurrent layer for reducing the complexity of the network. DCD-GAN \cite{Rain02_2022} is an effective unpaired image deraining adversarial framework which explores the mutual properties of the unpaired exemplars.
DRSformer\cite{chen2023learning} proposes an effective DeRaining network, Sparse Transformer that can adaptively keep the most useful self-attention values for feature aggregation so that the aggregated features better facilitate high-quality image reconstruction.
For desnowing, Wavelet~\cite{Snow01_2021} proposes a complex wavelet loss and a hierarchical decomposition paradigm to deal with snowflake removal of different sizes.
DesnowGAN\cite{Snow02_2020} employs a pyramidal hierarchical design to enrich location information and reduce computational complexity. All these single adverse weather removal approaches are specifically designed for one task, usually based on their corresponding physical models.

\textbf{All-in-one Adverse Weather Removal.} All-in-one methods aim to train a unified model that can tackle multiple types of adverse weather, e.g., haze, snow, and rain, simultaneously. Representative works include: GeMT~\cite{Multitask03_2020_CVPR}, which introduces the concept of all-in-one adverse weather removal and designs a generator with multiple task-specific encoders; MutiTS~\cite{Multitask01_2022_CVPR}, which proposes a two-stage knowledge learning mechanism based on a multi-teacher and student architecture; IPT model~\cite{Multitask02_2021_CVPR}, which is trained with multi-task data with multi-heads and multi-tails.
TransWeather\cite{TransWeather_2022_CVPR} trains a single encoder and single decoder transformer network to tackle all adverse weather removal problems, which adopts the weather type queries to handle the all-in-one problem at once.
SmartAssign\cite{wang2023smartassign}, a multi-task learning strategy, introduces a novel knowledge assignment approach to adaptively determine the shared and exclusively-used knowledge atoms for deraining and desnowing.
Among them, IPT\cite{Multitask02_2021_CVPR} and GeMT~\cite{Multitask03_2020_CVPR} exhibit continual network expansions with an increasing number of tasks and require the task ID during inference. 
MutiTS~\cite{Multitask01_2022_CVPR} requires multiple pre-trained models which will increase the training cost. 
TransWeather\cite{TransWeather_2022_CVPR} and SmartAssign\cite{wang2023smartassign} train the all-in-one model under the condition that all types of adverse weather removal training datasets need to be finely collected before model training.
However, when a new task emerges, these methods require to retrain the network, and as task data grows, the cost of each retraining also increases.
We aim to overcome the aforementioned challenges by employing incremental learning, which leverages the knowledge acquired from previous tasks to avoid resource wastage.

\subsection{Continual Learning Methods}

Existing incremental learning methods can be roughly divided into three categories: architectural-based, rehearsal-based, and regularization-based methods.

The architectural-based methods \cite{Art01_2016, Art02_2021, Art03_2022} dynamically expand the capacity of the network when training on a new task to overcome catastrophic forgetting in continual learning.
Progressive neural networks \cite{Art01_2016} extend each layer and add lateral connections between duplicates whenever learning a new task.
DER \cite{Art02_2021} proposes a novel two-stage learning approach that utilizes a dynamically expandable representation for more effective incremental concept modeling.
DyTox \cite{Art03_2022} adds task-specific tokens to achieve task-specialized embeddings for continual learning based on transformer architecture. 
However, most of these existing methods require a task ID during testing, and the parameters of the network will continue increasing.

Regularization-based methods can be divided into two categories. The first category focuses on constraining the change of parameters in the network during the learning process. Some studies, such as MAS\cite{MAS2018}, PIGWM \cite{PIGWM2021} and EWC \cite{EWC2017}, compute the importance of the parameters of a neural network for penalizing parameter changes. Some others, like  Adam-NSC \cite{Null_Space_2021_CVPR} and OWM \cite{Reg01_2019}, regulate the direction of parameter change. The second category is based on knowledge distillation, for example, LwF \cite{LwF2017} proposes to use knowledge distillation to prevent the representations of previous data from drifting too much when learning new tasks.

The rehearsal-based methods \cite{Rep01_2017, Rep02_2019, Rep03_2022,Reg_Rep01_2022} use a fixed-capacity buffer to keep a portion of old samples for replaying. 
iCaRL \cite{Rep01_2017} selects a small number of exemplars for replaying and nearest-mean-of-exemplar classification. 
BiC \cite{Rep02_2019} proposes a bias correction layer to solve the problem of data imbalance between old and new samples. 
GCR \cite{Rep03_2022} proposes gradient approximation as an optimization criterion for selecting `coresets', which closely approximates the gradient of all the data seen so far with respect to current model parameters.
Some methods~\cite{Reg_Rep01_2022, Reg_Rep02_2022, Reg_Rep03_2022, Podnet2020, AFC2022,clsparse} also combine parameter regularization with exemplars distillation for continual learning.

Recently, a few studies have explored incremental learning to train image restoration models~\cite{LIRA,PIGWM2021}. 
The most relevant work is PIGWM~\cite{PIGWM2021}, which proposes a parameter importance guided weight modification approach to overcome catastrophic forgetting for image de-raining. Compared with the task  PIGWM~\cite{PIGWM2021}, which only applies continual learning on a single adverse weather removal task, our task is more realistic and challenging, and faces severer catastrophic forgetting problem.
Overcoming catastrophic forgetting for multiple adverse weather conditions poses greater challenges.

\section{The Proposed Method}\label{sec:proposed method}
\subsection{Problem Setup}

In the setting of continual multiple adverse weather removal problem, the sequential training tasks can be denoted as $\{T_1, \cdots, T_t, \cdots, T_N\}$ of $N$ tasks, and each task includes its own corresponding type of adverse weather images.
Without losing generality, we study the problem with a simplified but representative setting with sequential input tasks $\{T_H, T_R, T_S \}$, 
which stands for image dehazing, deraining, and desnowing tasks, respectively. The corresponding datasets for these three tasks can be denoted as
$\mathcal{D}_H=\{\mathcal{I}_H, \mathcal{J}_H\}=\{({I}_i,{J}_i)\}_{i=1}^{N_H}$, $\mathcal{D}_R=\{\mathcal{I}_R, \mathcal{J}_R\}=\{({I}_i,{J}_i)\}_{i=1}^{N_R}$, and $\mathcal{D}_S=\{\mathcal{I}_S, \mathcal{J}_S\}=\{({I}_i,{J}_i)\}_{i=1}^{N_S}$, where ${I}_i$ is the adverse weather polluted image, ${J}_i$ is the corresponding ground-truth clean image, and $N_{t}$ is the number of training samples in the $t$-th task $T_t$. 

During the continual learning process, for the $t$-th task $T_t$, we are given the following three components: 1) The training dataset $\mathcal{D}_t$ for the $t$-th task $T_t$; 2) The previous model $\mathcal{F}_{\mathbf{w}}(\cdot,\mathbf{w}_{t-1})$ trained on previous dataset; 3) A tiny memory buffer \cite{ER} with fixed size which stores a very small portion of training data from all the previous tasks, denoted as $\mathcal{M}$, where our proposed method follows the knowledge replay based continual learning setting.


This task is very challenging for the following two reasons: 1) The model should contain knowledge for several adverse weather types simultaneously without accessing additional model costs; 2) The model should overcome severer \emph{catastrophic forgetting} problem during the continual learning process, as different types of adverse weather images often have their own intrinsic mapping rules, and the differences between them are always larger than that among different datasets within the same type of adverse weather as in PIGWM~\cite{PIGWM2021}. 

\subsection{Method Overview}
The proposed method aims to train a compact unified network architecture in a continual learning manner with sequential or alternative input training data, to tackle multiple adverse weather removal problems simultaneously. Here, we just consider three main weather removal tasks for simplicity, i.e., dehazing, deraining, and desnowing. 
Figure~\ref{fig::model} illustrates the overall framework with effective knowledge replay on a unified network structure, for continual all-in-one adverse weather removal tasks. The framework mainly includes three components: 1) the backbone network applied in the single weather image restoration task; 2) the effective knowledge replay based distillation on the network prediction; 3) the principal component projection in knowledge replay for better plasticity-stability trade-off in the continual learning.

\subsection{Training Scheme for Single Weather Removal }

Existing single weather removal algorithms have been well studied and have achieved promising performances~\cite{Haze01_2022, Snow01_2021, Rain02_2022}. In this paper, we adopt the representative FFA-Net~\cite{FFA2020} as our backbone network architecture, which is widely used in the image dehazing task. 
The FFA-Net~\cite{FFA2020} includes four components: the shallow feature extraction module, several group attention architectures, the feature concatenation module, and the reconstruction module with global residual skip connection.

As shown in Figure~\ref{fig::model}, the backbone network architecture can be divided into two parts: one is the feature extractor $\mathcal{F(\cdot)}$, which represents the output of the feature concatenation module;  another is the image projector head $\phi(\cdot)$ following the feature extractor $\mathcal{F(\cdot)}$. Denote the input bad weather polluted image as $I$, its corresponding restored clean image as $I_p=\phi(\mathcal{F}(I))$, and the ground-truth clean image $J$. To optimize the network parameters, we adopt the traditional $L_1$ loss and the contrast regularization as the reconstruction objective function, where the $L_1$ loss can be denoted as $\mathcal{L}_1=|\phi(\mathcal{F}(I)) -J|$, and the contrast regularization~\cite{Haze01_2022} can be denoted as follows,
\begin{equation}
    \begin{split}
        &\mathcal{L}_{CT}(\phi(\mathcal{F}(I)),J, I) = \\
        & \!-\! \sum_{l=1}^{L} w_l \log \! \frac{e^{\!-\!\left|E_{l}(\phi(\mathcal{F}(I)))\! - \! E_{l}(J) \right| / \tau}}{e^{\!-\!\left|E_{l}(\phi(\mathcal{F}(I)))\! -\! E_{l}(J) \right| / \tau}\! +\! e^{\!-\!\left|E_{l}(\phi(\mathcal{F}(I)))\! - \! E_{l}(I) \right| / \tau}},
    \end{split}
    \label{LCT}
\end{equation}
where ``$e$'' denotes the exponential operation, $E_l$ with $l=\{1,2,\cdots, L\}$ extracts the $l$-th hidden layer features from the fixed pre-trained model VGG-19~\cite{VGG_2014}, $\tau >0$ is the temperature parameter that controls the sharpness of the output, $|\cdot|$ denotes the $L_1$ distance, which usually works better than $L_2$ distance for image restoration task~\cite{zhao2015loss}, and $w_l$ is the weight coefficient for the $l$-th hidden feature from the fixed VGG-19 network. The configurations of parameter $w_l$ and $\tau$ follow \cite{Haze01_2022}. In Eqn.~\eqref{LCT}, we can see that $\phi(\mathcal{F}(I)),J, I$ serve as the \emph{anchor}, \emph{positive} and \emph{negative} point, respectively. Minimizing Eqn.~\eqref{LCT} encourages the restored image $\phi(\mathcal{F}(I))$ to be close to its corresponding ground-truth clean image $J$, while far away from the original bad weather polluted image $I$.

Then for the baseline/single weather image restoration task, the reconstruction objective function arrives at,  
\begin{equation}
    \begin{split}
        \mathcal{L}_{SW}= |\phi(\mathcal{F}(I)) -J| + \beta_{1} \mathcal{L}_{CT}(\phi(\mathcal{F}(I)),J, I),
    \end{split}
    \label{SingleWeatherLoss}
\end{equation}
where $\beta_{1}$ is the hyper-parameter to balance $\mathcal{L}_1$ and the contrast regularization term $\mathcal{L}_{CT}$. Note that, for all the baseline methods (i.e., each single adverse weather removal task, and fine-tuning settings), Eq.~\eqref{SingleWeatherLoss} is adopted as the objective function to train the backbone network.

\subsection{Effective Knowledge Replay on a Unified Network Structure}
To continually learn the unified adverse weather removal model through the sequential input data, we propose the effective knowledge replay based distillation framework, as illustrated in Figure~\ref{fig::model}. 
In order to maintain the old knowledge during continual learning, we have verified that the memory-based knowledge replay method is very effective to overcome severe \emph{catastrophic forgetting} for training such a unified adverse weather removal model.  Therefore, we design a fixed-size memory bank $\mathcal{M}$ to store a very small portion of training data from all previous tasks, to memorize previous knowledge. 
Relying on memory $\mathcal{M}$, we propose a novel and effective knowledge replay method that replays the learned knowledge in the network with a principal component projection on the mid-level representation (as discussed in Sec. \ref{sec:proj}) and the network response on the old task samples (as discussed in the following). 

Specifically, to train the $t$-th ($t \ge 2$) adverse weather removal task, we are given: 1) current training dataset $\mathcal{D}_t=\{(I_i^{t}, J_i^{t})\}_{i=1}^{N_t}$; 2) some previous task data stored in the memory bank $\mathcal{M}= \{\bar{I}_i^{t-1}\}_{i=1}^{N_m} = \{\{I_{i}^1\}_{i=1}^{N_m/(t-1)}, \cdots,\{I_{i}^{t-1}\}_{i=1}^{N_m/(t-1)}\}$, where $N_m$ is the size of the memory bank; 3) the previously trained model denoted as $\phi^{t-1}(\mathcal{F}^{t-1})$. Note that, the data stored in the memory bank requires no ground-truth clean images, which is the main difference from the traditional rehearsal-based class incremental learning methods~\cite{Rep01_2017, Rep02_2019, Rep03_2022}. The proposed method just utilizes the previous data without references to memorize effective knowledge to alleviate \emph{catastrophic forgetting}. Such a setting without ground-truth labels consumes a relatively small cost and is friendly for model training, as it saves half of the memory space compared to some methods storing the pair-wise training samples. Extensive experiments have also illustrated its effectiveness by performing knowledge distillation with predicted soft labels, where such findings have also been demonstrated in~\cite{yuan2020revisiting}.

As illustrated in Figure~\ref{fig::model}, current task data $\mathcal{D}_t$ passes through the current new model $\phi^t(\mathcal{F}^{t})$, where $\phi^t(\mathcal{F}^{t})$ is initialized by the previous model $\phi^{t-1}(\mathcal{F}^{t-1})$. Then, the objective function $\mathcal{L}_{SW}$ shown in Eq.~\eqref{SingleWeatherLoss} is adopted to optimized current model $\phi^t(\mathcal{F}^{t})$. Meanwhile, these non-reference images stored in $\mathcal{M}$ act as the knowledge replay to memorize previous knowledge. Specifically, the image $\bar{I}_i^{t-1}$ passes through both of the old model $\phi^{t-1}(\mathcal{F}^{t-1})$ and current model $\phi^t(\mathcal{F}^{t})$, where $\phi^{t-1}(\mathcal{F}^{t-1})$ is fixed during the model distillation training. Therefore, the knowledge replay distillation loss can be expressed as,  
\begin{equation}
    \begin{split}
        &\mathcal{L}_{KD} = |\phi^{t-1}(\mathcal{F}^{t-1}(\bar{I}_i^{t-1})) - \phi^{t}(\mathcal{F}^{t}(\bar{I}_i^{t-1}))| \\ 
        &+ \beta_{2} \mathcal{L}_{CT}(\phi^{t}(\mathcal{F}^{t}(\bar{I}_i^{t-1})), \phi^{t-1}(\mathcal{F}^{t-1}(\bar{I}_i^{t-1})), \bar{I}_i^{t-1}),
    \end{split}
    \label{KDLoss}
\end{equation}
where $\beta_2$ is a hyper-parameter to balance $\mathcal{L}_1$ and $\mathcal{L}_{CT}$. Eq.~\eqref{KDLoss} shows that, both $\mathcal{L}_1$ loss and the contrast regularization $\mathcal{L}_{CT}$ are involved in the knowledge distillation framework, which performs only on the images without references of previous tasks in $\mathcal{M}$. We have thoroughly verified that such a kind of knowledge distillation strategy works better than that of simply using the experience replay method~\cite{ER}, which directly uses the ground truth of previous and current tasks for joint fine-tuning. This shows that the old model is very helpful to memorize some useful knowledge of previous tasks.

\subsection{Effective Knowledge Replay with Principal Component Projection}
\label{sec:proj}


\begin{figure}[h] 
  \centering
   \includegraphics[width=0.5\textwidth]{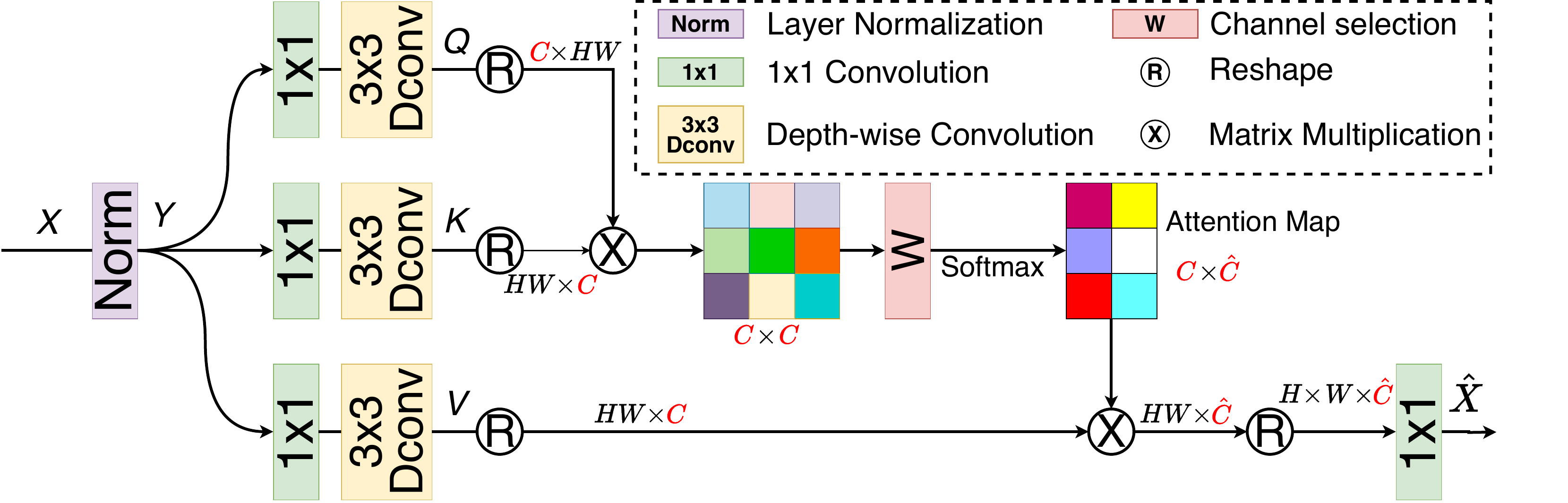}
   \caption{ Detail architecture of the principal component projection. } 
   \label{fig::autoencoder} 
\end{figure}

The adverse weather image restoration can be considered as a many-to-one feature mapping problem. That is to say, image features from different adverse weather domains ($e.g.$, hazy, rain, and snow) are first transformed into the clean image features through the clean feature extractor $\mathcal{F}(\cdot)$, and then the clean image features are transformed into the clean natural image through the clean image projector $\phi(\cdot)$. Thus, the key knowledge lies in the clean feature extractor $\mathcal{F}(\cdot)$. Then we further propose to perform knowledge distillation on the middle-level clean feature obtained from $\mathcal{F}$, to achieve model stability on previous tasks.    

To achieve a better plasticity-stability trade-off, we propose to preserve and transfer principal knowledge in the clean feature space. Generally, PCA~\cite{PCA_1901} is commonly used for dimension reduction by projecting the data onto the lower dimension space, which preserves most of the variation in the data. However, it usually assumes huge computation costs or is even impossible to perform PCA on these extremely high-dimension data, that is, intermediate features of the deep neural network. Fortunately,  existing auto-encoder techniques could learn an efficient representation/encoding for a set of data in a self-supervised manner, typically for dimension reduction. 
Specifically, we first feed the images stored in the memory bank as input to the newly trained network, and obtain the corresponding intermediate layer features. Subsequently, the obtained features can be used to train an autoencoder, and the obtained encoder can be utilized as a projector to extract principal components of intermediate layer features for the following distillation training.
Finally, we verify that the efficient encoding obtained by the encoder could mimic PCA and capture the principal components of the input features. 

We use the auto-encoder framework to mimic the principal component projection, where the network architecture of the encoder can refer to~\cite{restormer2022}, while the decoder is implemented by two convolution layers of kernel size $3\times 3$. As illustrated in Figure\ref{fig::autoencoder}, given an input feature tensor $\mathbf{X}\in \mathbb{R}^{H\times W\times C}$, which is normalized to obtain $\textbf{Y}\in \mathbb{R}^{H\times W\times C}$, our multi-head transposed attention (MHTA) first generates the query (Q), key (K) and value (V) projections~\cite{restormer2022}. They are achieved by applying the $1\times 1$ convolutions to aggregate pixel-wise cross-channel context, followed by $3\times 3$ depth-wise convolution to encode the spatial context. After that, we perform tensor reshape in the spatial dimension for all Q, K and V, and then obtain the channel attention map $A\in \mathbb{R}^{C\times C}$ by $Q\cdot K/\tau$, where $\tau$ is a hyper-parameter. In the following, we perform the channel selection block $\mathcal{W}$ to reduce the dimension. Specifically, we apply a learnable matrix $\mathcal{W}\in \mathbb{R}^{C\times \hat{C}}$ to  project the original attention $A$ to a lower dimension through the following operation: $\hat{A}=Softmax(A \cdot \mathcal{W})\in \mathbb{R}^{C\times \hat{C}}$, where $Softmax(\cdot)$ is the normalization operation, and $\hat{C}$ is smaller than $C$. Next, we obtain the final output $\hat{X}\in \mathbb{R}^{H\times W\times \hat{C}}$ by $V\hat{A}$ following the spatial reshaping and $1\times 1$ convolution. In the principal component projection module, the main reason for adopting the channel attention instead of the spatial attention is the computation overhead. As we know, the major computation overhead in transformers comes from the self-attention layer. In the traditional spatial attention, the time and memory complexity of the key-query dot product interaction grows quadratically with the spatial resolution of input, $i.e.$, $\mathcal{O}(W^2H^2)$ for images of size $W\times H$~\cite{restormer2022}. Therefore, it is infeasible to apply such spatial attention to most image restoration tasks which often involve high-resolution images. To alleviate this issue, we propose the channel attention followed by the channel selection block, which has linear complexity, $i.e.$, $\mathcal{O}(C^2)$, where the size of the channel is usually relatively smaller.

The auto-encoder is trained on the image data stored in the memory bank in a self-supervised manner since the knowledge distillation only performs on these samples during continual learning. 
It is worth noting that the parameters of the auto-encoder are trained offline, and are frozen during backbone network training.
Given the image $\bar{I}_i^{t-1}$ and the previous fixed clean feature extractor $\mathcal{F}^{t-1}(\cdot)$, we then extract the features for all the images stored in $\mathcal{M}$, denoted as $\mathcal{D}_F=\{\mathcal{F}^{t-1}(\bar{I}_i^{t-1})\}_{i=1}^{N_m}$.  The feature set $\mathcal{D}_F$ acts as the training samples to optimize the auto-encoder network architecture $\Phi(\cdot)$, and $L1$ loss $\mathcal{L}_1=|\mathcal{F}^{t-1}(\bar{I}_i^{t-1}) - \Phi(\mathcal{F}^{t-1}(\bar{I}_i^{t-1}))|$ is adopted as the objective function. Note that, the auto-encoder network $\Phi(\cdot)$ is trained before every new task participating in the continual training, as the memory bank is updated at the end of each training task. At this time, samples for every task are balanced, and this balanced distribution prevents task bias from occurring in the autoencoder. Then, we optimize the adverse weather removal model under the knowledge replay based distillation framework with a fixed encoder in $\Phi(\cdot)$ within the current task.  

Define the encoder in $\Phi(\cdot)$ as $\psi(\cdot)$, which works as the principal component projection module in the knowledge distillation based continual learning. Specifically, during the continual learning, the image $\bar{I}_i^{t-1}$ in the memory bank passes through the previous fixed clean feature extractor $\mathcal{F}^{t-1}$ and current feature extractor $\mathcal{F}^{t}$, separately. Following the shared principal component projection module $\psi(\cdot)$, we can obtain their corresponding efficient encodings $\psi(\mathcal{F}^{t-1}(\bar{I}_i^{t-1}))$ and $\psi(\mathcal{F}^{t}(\bar{I}_i^{t-1}))$. Therefore, the principal knowledge distillation can be expressed as follows,
\begin{equation}
   \mathcal{L}_{PKD} =  \left | \psi(\mathcal{F}^{t-1}(\bar{I}_i^{t-1})) - \psi(\mathcal{F}^{t}(\bar{I}_i^{t-1})) \right |,
    \label{ploss}
\end{equation}
By applying this principle knowledge distillation objective function, it helps to improve the plasticity of the network while alleviating \emph{catastrophic forgetting}.

\subsection{The Overall Objective Function}

In the proposed CL framework, at each stage with the current weather removal task and the memory buffer (with samples from old tasks), the model is updated by optimizing the following overall objective function: 
\begin{equation}
    \mathcal{L} = \mathcal{L}_{SW} + \alpha\mathcal{L}_{KD} + \lambda\mathcal{L}_{PKD} ,
    \label{overallLoss}
\end{equation}
where $\alpha$ and $\lambda$ are two hyper-parameters to balance the above three terms in the overall objective function. Note that, the first term $\mathcal{L}_{SW}$ is calculated with the samples from the current task, and the following two terms $\mathcal{L}_{KD}$ and $\mathcal{L}_{PKD}$  are calculated with the old-task samples stored in the tiny memory buffer.

\section{Experiments}\label{sec:experiments}

\subsection{Dataset and Experiment Setting}


The following three datasets are adopted in the proposed continual all-in-one adverse weather removal task, i.e., OTS~\cite{OTS2019}, Rain100H \cite{Rain100H_2017_CVPR}, and Snow100K~\cite{Snow100K2018}, corresponds to one type of adverse weather. OTS is the subset of the RESIDE dataset, which consists of 72,135 synthetic hazy images for training and 500 images for testing. Rain100H has 1800 rainy images for training and 200 for testing. Snow100K has 50,000 synthetic snow images for training, and we test on Snow100k-M (16,588 snowy images). While for OTS and Snow100K, we randomly crop the images to a size of 240x240 as input. In order to maintain comparable dataset sizes for joint training among three datasets, we perform upsampling on 1,800 images from Rain100H to obtain 72,000 images, and the patch size is 240 with a stride of 25. We also tested the performance of our model on real-world images of adverse weather conditions. The subset "Unannotated Real-world Haze images" of RESIDE contains 4,809 real images with haze, and Snow100K provides 1,329 real snowy images. In order to evaluate the real-world rain removal performance, we employed the test dataset provided by Real200\cite{real200}, containing 200 real rainy images.


We trained the model in an incremental learning manner, following the task order of image dehazing$\rightarrow$deraining$\rightarrow$desnowing. We apply the peak signal-to-noise ratio (PSNR) and the structural similarity (SSIM) for quantitative evaluation, which are widely used to evaluate image quality for image restoration tasks. We finally report the \emph{Average} metrics of PSNR and SSIM on all the tasks for the continual learning experimental setting.

\subsection{Implementation Details}
We implement the proposed method based on PyTorch with NVIDIA GTX 3090 GPUs. The optimization configuration is set according to \cite{Haze01_2022}, with an initial learning rate of 0.0001 and adopting the cosine annealing strategy. We adopt the Adam optimizer with a default exponential decay rate of 0.9. We empirically set $\beta_{1}$ in Eq.~\eqref{SingleWeatherLoss} as 0.8, $\beta_{2}$ in Eqn.~\eqref{KDLoss} as 0.2, and the values of parameters $\alpha$ and $\lambda$ in Eq. \eqref{overallLoss} to 1.0 and 0.3, respectively. Following \cite{Haze01_2022}, hyper-parameter $\tau$ in Eq. \eqref{LCT} is set to 0.5, while $w_l$'s are set as $\frac{1}{32}$, $\frac{1}{16}$, $\frac{1}{8}$, $\frac{1}{4}$ and 1, respectively. The size of the memory buffer is fixed as 500, which is less than $1\%$ of the entire dataset. The samples are randomly selected, and the memory size allocated for each task is the same. To ensure a fair comparison, after randomly selecting the samples, all compared methods use the same set of memory exemplars. We train the model with batch size 1 for both new data and old exemplars. The feature maps extracted by the feature extractor have 192 channels, while we set the output channel number of the principal projector as 16.

\begin{table*}[]
	\footnotesize
	\centering
        \caption{Comparison with representative CL methods under the setting of dehazing $\rightarrow$ deraining $\rightarrow$ desnowing.}
	\vspace{0pt}
	\scalebox{1}[1]{
		\begin{tabular}{ccccccccc}
			\toprule
			\multirow{2}{*}{\centering Methods} & \multicolumn{2}{c}{Average} & \multicolumn{2}{c}{Task1-OTS} & \multicolumn{2}{c}{Task2-Rain100H} & \multicolumn{2}{c}{Task3-Snow100K} \\
			\cmidrule{2-9}  & PSNR & SSIM & PSNR & SSIM & PSNR & SSIM & PSNR & SSIM\\
			\midrule	
            Individual-Training                          & 31.69 & 0.9444 & 31.44 & 0.9795 & 29.47 & 0.9072 & 34.17 & 0.9464 \\
            Joint-Training                               & 31.27 & 0.9405 & 31.27 & 0.9780 & 29.48 & 0.9059 & 33.07 & 0.9377 \\
            Fine-Tuning                                    & 20.08 & 0.7116 & 16.24 & 0.8008 & 13.83 & 0.4292 & 30.17 & 0.9047 \\
            Joint-M &24.94 	&0.5838 	&15.93 	&0.8106 	&25.10 	&0.8072 	    &33.80 	&0.9436   \\
                \midrule
			EWC\cite{EWC2017}                       & 22.26 & 0.7192 & 28.33 & 0.9577 & 13.74 & 0.4024 & 24.73 & 0.7975 \\
			MAS\cite{MAS2018}                       & 23.03 & 0.7489 & 29.18 & 0.9661 & 16.68 & 0.5010 & 23.23 & 0.7797 \\
                LwF\cite{LwF2017}                       & 19.32 & 0.6803 & 20.87 & 0.8544 & 13.03 & 0.3763 & 24.05 & 0.8101 \\
			POD\cite{Podnet2020}                    & 21.30 & 0.6988 & 20.49 & 0.8296 & 15.35 & 0.3988 & 28.08 & 0.8680 \\
            PIGWM\cite{PIGWM2021}                   & 22.46 	&0.7214 	&28.93 	&0.9522 	&14.01 	&0.4162 	&24.45 	&0.7959  \\
            AFC\cite{AFC2022}                       & 26.27 & 0.8164 & 26.99 & 0.9012 & 22.48 & 0.6567 & 29.34 & 0.8911 \\

                \midrule
            TransWeather\cite{TransWeather_2022_CVPR}                     & 28.56 & 0.9064 & 28.52 & 0.9612 & 25.20 & 0.8333 & 31.95 & 0.9247 \\
            MutiTS\cite{Multitask01_2022_CVPR}             & 29.87 & 0.9237 & 29.96 & 0.9692 & 26.06 & 0.8606 & 33.59 & 0.9412 \\
            
                \midrule
            ER-$\mathcal{L}_{SW}$                     &29.67 &0.9249  &29.09 &0.9693  &27.38 &0.8718  &32.55 &0.9338  \\
            $\mathcal{L}_{SW}$ + $\mathcal{L}_{KD}$                & 30.36    & 0.9339   & 29.66 & 0.9717  & 28.54 & 0.8934 & 32.90 & 0.9366  \\
            $\mathcal{L}_{SW} + \mathcal{L}_{KD} + \mathcal{L}_{PKD}$    & \textbf{30.70} & \textbf{0.9335} & 31.03 & 0.9775 & 28.52 & 0.8900 & 32.54 & 0.9328 \\
            
                \bottomrule
		\end{tabular} 
  }
		\label{table:Three task}
\end{table*}

\begin{table}[]
    \scriptsize
    \centering
    \caption{Quantitative comparisons on the setting of dehazing $\rightarrow$ deraining.}
    \setlength\tabcolsep{3.5pt}
    \scalebox{1}[1]{
    \begin{tabular}{ccccccc}
        \toprule
        \multirow{2}{*}{\centering Methods}
        
        & \multicolumn{2}{c}{Average}
        & \multicolumn{2}{c}{Task1-OTS}
        & \multicolumn{2}{c}{Task2-Rain100H}
        \\
        \cmidrule{2-7} & PSNR & SSIM  & PSNR & SSIM  & PSNR & SSIM \\
        \midrule
        Individual-Training         & 30.46 & 0.9434 & 31.44 & 0.9795 & 29.47 & 0.9072          \\
        Joint-Training              &30.31 &0.9412   &31.30 	&0.9782 	&29.32 &0.9041        \\
        Fine-Tuning                         & 22.66 & 0.8502    & 16.24 & 0.7998  & 29.08 & 0.9006  \\
        Joint-M                         & 22.62 & 0.7183    & 16.03 & 0.5344  & 29.21 & 0.9023  \\
            \midrule
        EWC \cite{EWC2017}                & 25.19 & 0.8223    & 28.51 & 0.9594  & 21.87 & 0.6852  \\
        MAS \cite{MAS2018}                & 24.77    & 0.7987   & 28.41 & 0.9604  & 21.13 & 0.6369   \\
            LwF \cite{LwF2017}                & 18.44    & 0.6349   & 23.43 & 0.8882  & 12.45 & 0.3816   \\
        POD \cite{Podnet2020}             & 22.26    & 0.7637   & 21.66 & 0.7965  & 22.86 & 0.7309   \\
        PIGWM \cite{PIGWM2021}            & 25.24 & 0.8294    & 27.88 & 0.9440  & 22.61 & 0.7149   \\
        AFC \cite{AFC2022}                & 26.56    & 0.8712   & 27.75 & 0.9281  & 25.38 & 0.8144   \\
        \midrule
        ER-$\mathcal{L}_{SW}$                       & 28.97    & 0.9322   & 29.12 & 0.9701  & 28.82 & 0.8942 \\
        $\mathcal{L}_{SW} + \mathcal{L}_{KD}$                & 29.21    & 0.9349   & 29.54 & 0.9724  & 28.89 & 0.8975  \\
        $\mathcal{L}_{SW} + \mathcal{L}_{KD} + \mathcal{L}_{PKD}$         & \textbf{29.99} & \textbf{0.9348}  & 31.35 & 0.9792  & 28.64 & 0.8903  \\
        \bottomrule
    \end{tabular}}
    \label{table: two task}
\end{table}

\begin{figure*}[!htbp]
\centering
\subfloat[Input]{
    \includegraphics[width=1.66in]{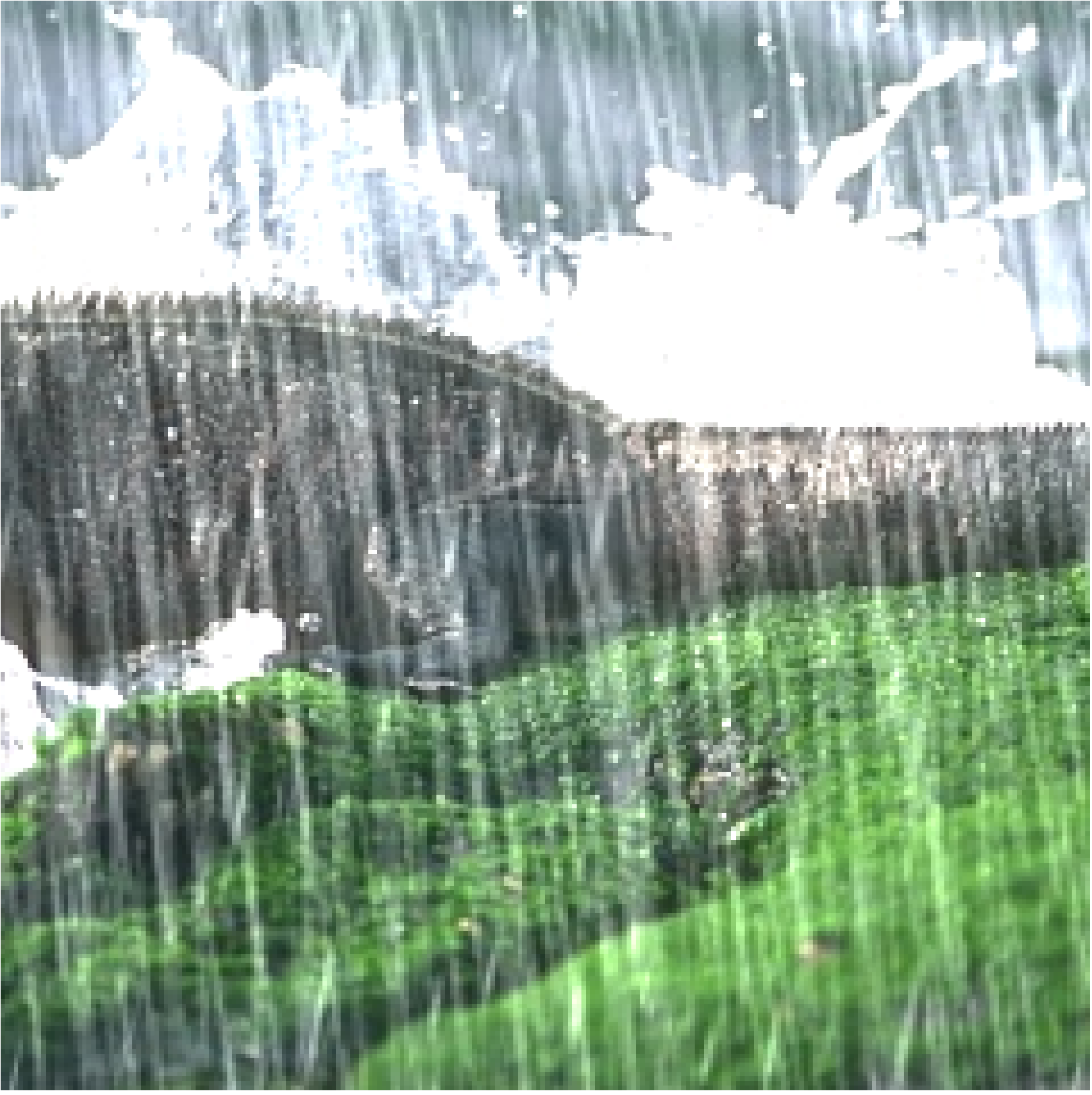}
    }
\hspace{0\textwidth}
\subfloat[Feature map extracted by feature extractor $\mathcal{F}$]{
    \includegraphics[width=3.32in]{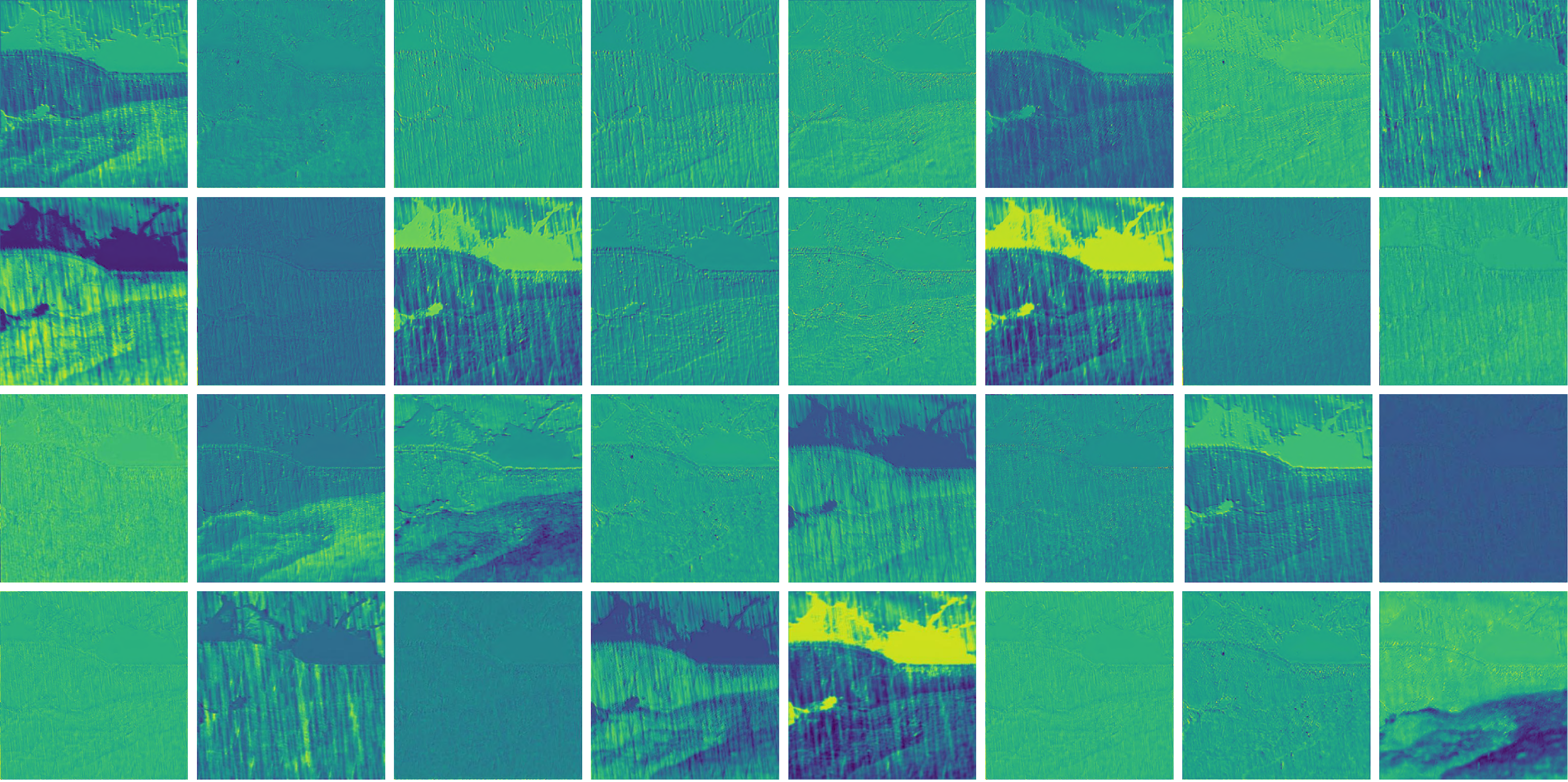}
    }
\hspace{0\textwidth}
\subfloat[Principal Component Projection]{
    \footnotesize
    \includegraphics[width=1.66in]{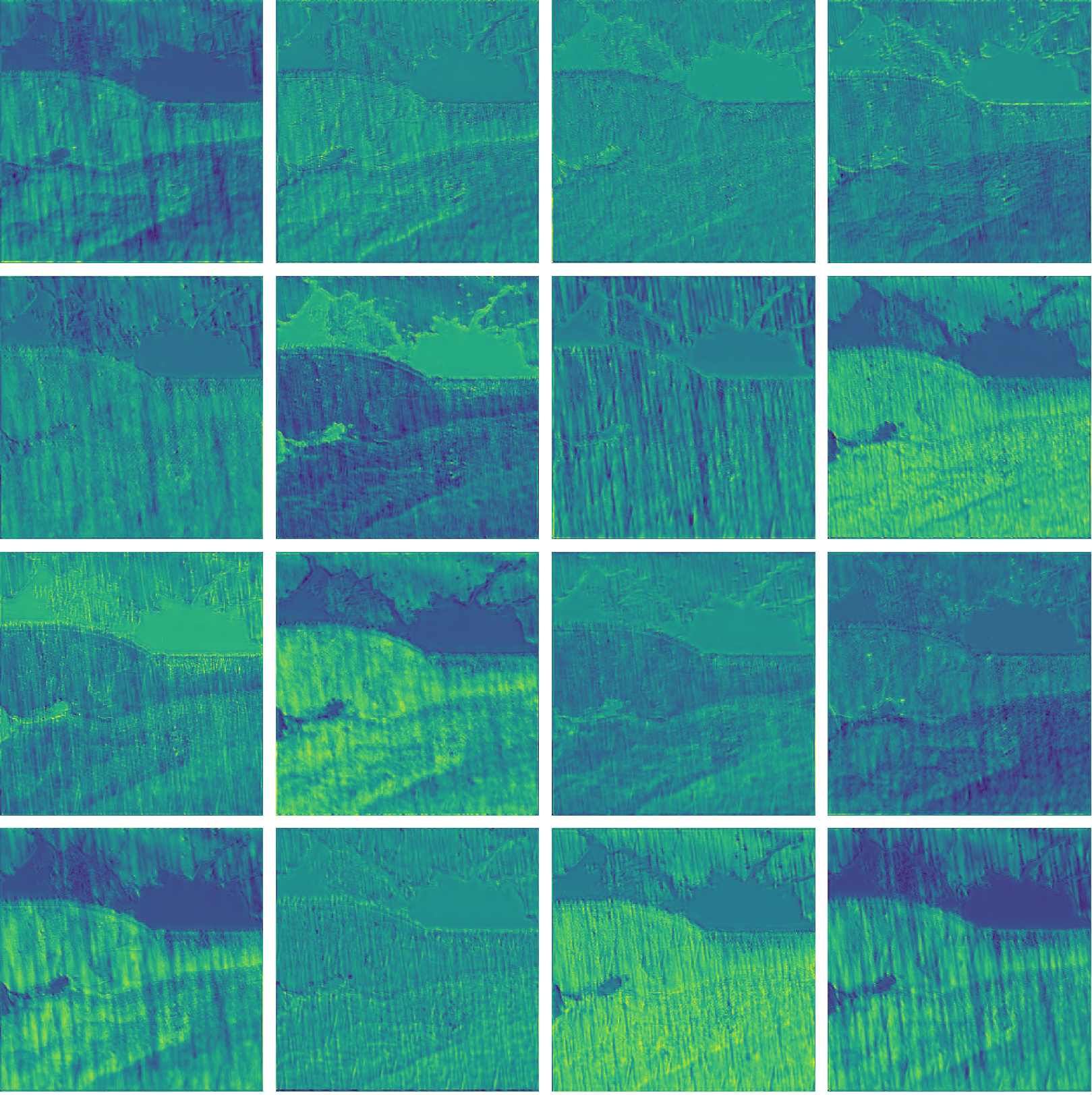}
    }
\caption{Feature map visualization of intermediate features extracted by $\mathcal{F}$ in (b), and the principal component encoder $\psi(\mathcal{F})$ in (c).}
\label{fig::h} 
\end{figure*}

\begin{figure*}[htbp] 
    \centering 
    \includegraphics[width=\textwidth]{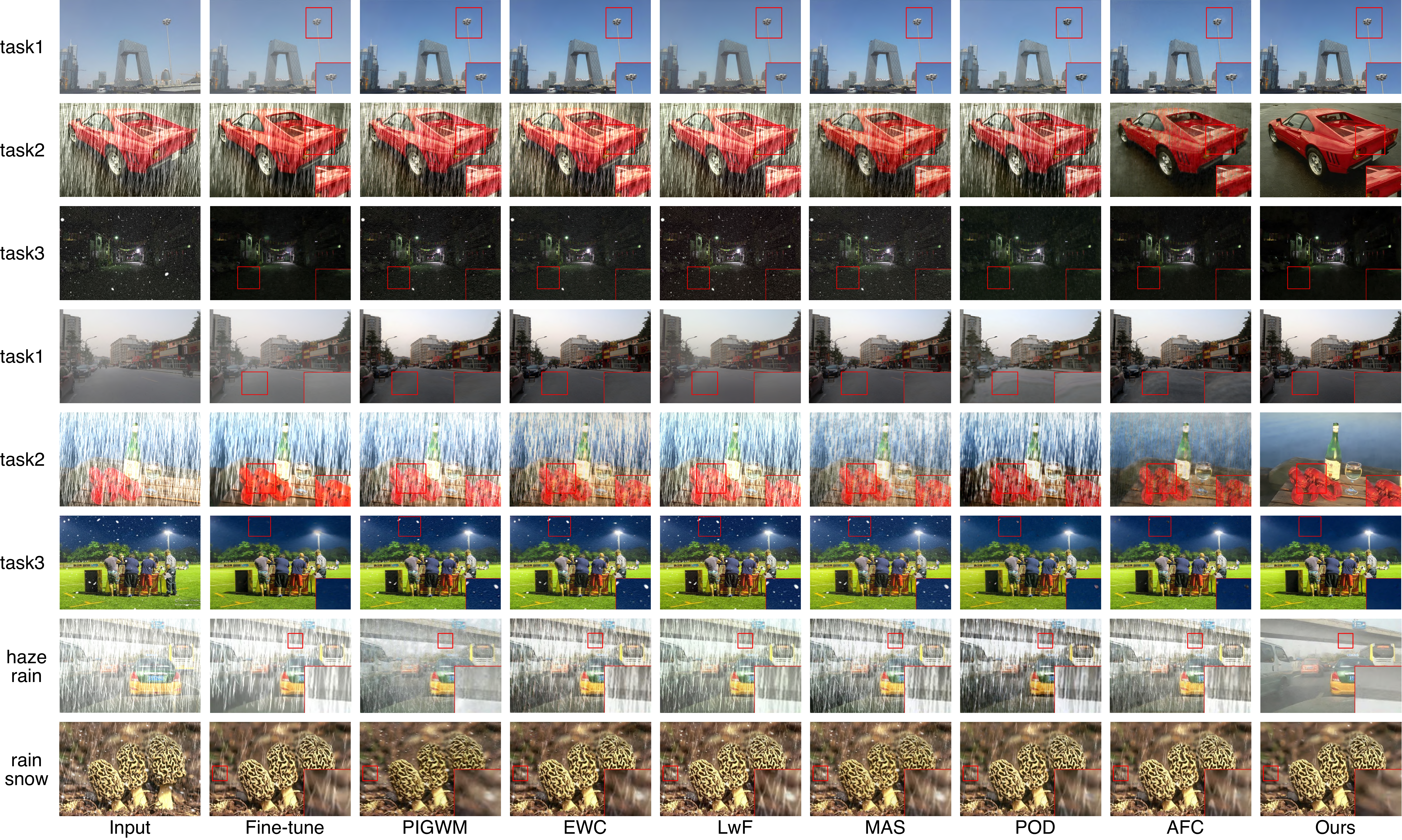}
    \captionsetup{justification=centering}
    \caption{ Visualization comparison of adverse weather removal using different continual learning algorithms. } 
    \label{fig::visual}
    \vspace{-10pt}
\end{figure*}

\begin{figure*}[htbp] 
    \centering 
    \includegraphics[width=\textwidth]{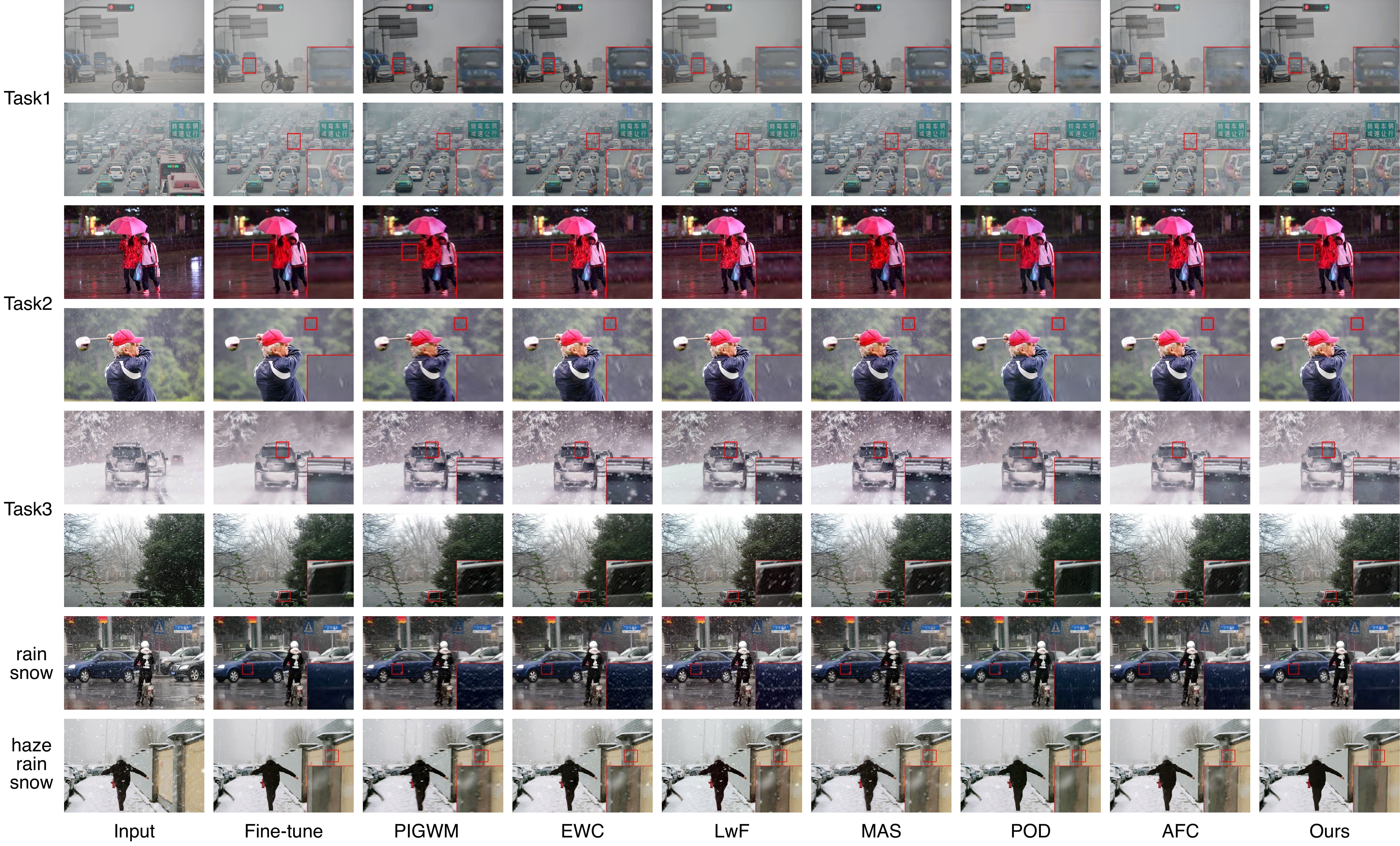} 
    \caption{  Visual quality comparison on real-world degraded images, under hazy, rainy and snow scenarios. } 
    \label{fig::real-visual}
    \vspace{-10pt}
\end{figure*}

\subsection{Quantitative Evaluation}
In this section, we evaluate the proposed method for adverse weather removal task on two sequential settings: dehazing $\rightarrow$ deraining of two tasks, and dehazing $\rightarrow$ deraining $\rightarrow$ desnowing of three tasks. Since we are the first to propose continual learning on multiple adverse weather removal tasks with unified network architecture, we implement the following two sets of methods to extensively illustrate the effectiveness of the proposed method. 
The first set is four baseline methods: 1) Individual-Training, which trains and tests the network model on an individual dataset with the objective function $\mathcal{L}_{SW}$; 2) Joint-Training, which trains the network model on all the training samples of these three tasks jointly, with objective function $\mathcal{L}_{SW}$; 3) Fine-Tuning, which trains the network model sequentially with the three datasets by fine-tuning strategy; 4) Joint-M, which uses data in the memory bank together with the new data for joint-training, with objective function $\mathcal{L}_{SW}$. Note that, for Joint-Training, Joint-M, and Fine-Tuning methods, we adopt the final unified model to test on these three testing datasets. The ``Individual-Training'' and ``Joint-Training'' could work as the upper bound under this experiment setting. 
Secondly, we also compare the proposed method with six representative continual learning methods, i.e., EWC\cite{EWC2017}, LwF\cite{LwF2017}, MAS\cite{MAS2018}, POD\cite{Podnet2020}, PIGWM\cite{PIGWM2021} and  AFC\cite{AFC2022}. We implement these methods based on their publicly released codes with the same network architecture as ours for fair comparisons.

Table~\ref{table:Three task} and Table~\ref{table: two task} show the evaluation results on the settings of dehazing $\rightarrow$ deraining $\rightarrow$ desnowing and dehazing $\rightarrow$ deraining, respectively. Compared with the baseline methods, we can clearly see that the proposed method improves the baseline method (Fine-tuning) a lot, and is just $0.99$db PSNR lower than the Individual-Training. When compared with other representative works, the proposed method achieves the best performance. It shows that our method can achieve a better stability-plasticity trade-off than other methods among these three tasks, which greatly illustrates the effectiveness of the proposed framework. The visualization results are shown in Figure \ref{fig::visual}. TransWeather and MutiTS are two all-in-one adverse weather removal approaches that simultaneously utilize all available data, and they should have a better performance. However, the performance of these two methods is not satisfactory because they exhibit poor results on the Rain100H dataset, particularly when it comes to removing dense rain streaks. Joint-Training, which trains the network model on all the training samples of these three tasks jointly, working as our upper bound under this experiment setting.
We also conduct experiments of the Joint-M, where the model is just trained with the old data from the memory bank and current new data together. The results are shown in Table \ref{table:Three task} and Table \ref{table: two task}, we can see that such joint training is a little better than the baseline method Fine-tuning in the experiments involving the three tasks, but far behind ours. One of the reasons is that such a joint training strategy faces severe imbalance in the number of training samples between new and old tasks. As a result, during such a joint-training strategy, the model tends to heavily favor the new task. On the contrary, our distillation training strategy provides an effective way to preserve knowledge learned from the old tasks.

As illustrated in Eq.~\eqref{overallLoss}, the proposed method contains the basic reconstruction loss $\mathcal{L}_{SW}$, the knowledge distillation loss on the network prediction $\mathcal{L}_{KD}$ and principal feature distillation loss $\mathcal{L}_{PKD}$. To reveal how each term contributes to the performance improvement, we implement the following three variants of the proposed method as shown in Table~\ref{table:Three task} and Table~\ref{table: two task}: 1) ER-$\mathcal{L}_{SW}$, which acts as one strong baseline method, and uses current training samples as well as the training images stored in the memory bank $\mathcal{M}$ with references, to train network model with the objective function $\mathcal{L}_{SW}$; 2) $\mathcal{L}_{SW}$+$\mathcal{L}_{KD}$, which trains the network under the knowledge distillation framework, with training loss $\mathcal{L}_{SW}$+$\mathcal{L}_{KD}$; 3) $\mathcal{L}_{SW}$+$\mathcal{L}_{KD}$+$\mathcal{L}_{PKD}$, which trains the network with the whole objective function as illustrated in Eqn.~\eqref{overallLoss}. Note that, only training the baseline method ER-$\mathcal{L}_{SW}$ requires storing the pairwise bad weather polluted and ground-truth clean images in $\mathcal{M}$, the other two methods just need to store the previous bad weather polluted images in $\mathcal{M}$, which is more efficient and practical.

Experiment results in Table~\ref{table:Three task} and Table~\ref{table: two task} show that, $\mathcal{L}_{SW}$+$\mathcal{L}_{KD}$ outperforms the baseline ER-$\mathcal{L}_{SW}$ method by a margin of $0.69$db and $0.24$db PSNR on the three-task and two-task settings, respectively. The only difference between them is knowledge replay by ground-truth clean image or the old model prediction. This implies that the old model itself also contains a significant amount of information, which also makes the training smoother and easier in practice. When further applying the principal knowledge distillation $\mathcal{L}_{PKD}$ on the intermediate features, another $0.34$db and  $0.78$db PSNR performance gains can be obtained on these two settings, which greatly illustrate the effectiveness of the proposed method.

Besides, even though we train the network in an incremental learning manner, our approach closely approaches the upper bound. At the end of the second task, the network's dehazing performance only decreases by 0.09db PSNR, and at the end of the third task, the deraining performance only drops by 0.12db PSNR. Our proposed method exhibits a low forgetting rate, which demonstrates its strong robustness for long sequences.

\textbf{Differences from All-in-One Adverse Weather Removal Methods:} The all-in-one framework utilizes a single unified model to address all adverse weather removal problems. In the following, we present the differences between our method and existing all-in-one approaches. Firstly, and most importantly, the all-in-one methods require the use of all available data during training, which poses a challenge when dealing with more realistic scenarios where data arrives incrementally. Only our method can handle such scenarios because it mitigates the catastrophic forgetting issue commonly encountered in incremental learning. Secondly, training existing all-in-one models requires storing data for all tasks, while our method only requires to store data for one task at a time and utilizes a very small memory buffer to store previous task data. Thirdly, when a new task emerges, the existing all-in-one model needs to be retrained, leading to wasted resources from previous training. In contrast, our method can continue to leverage the previously trained model.

\subsection{Qualitative Evaluation}
\textbf{Synthetic Images. }We train the network using different methods in an incremental learning manner in the order of dehazing, deraining, and desnowing, and perform visual tests with the final model obtained. According to Figure \ref{fig::visual}, we conduct visual tests on the models using two sets of images. Due to the fact that finetuning does not retain the old knowledge, \textit{catastrophic forgetting} occurs, resulting in the loss of the ability to perform dehazing and deraining tasks, and the model is only capable of handling the desnowing task. The majority of the methods preserve some dehazing capability, but encounter difficulties in preserving the ability to perform deraining. Our method achieves superior performance with less distortion in the recovered images compared to AFC~\cite{AFC2022}. 
In Figure \ref{fig::visual}, we also demonstrate scenarios with combined degradations, such as haze and rain, rain and snow, where our method also achieves the best results. We also observe some limitations of our trained model. When the haze concentration is high in scenes with haze and rain, the image clarity is reduced after the restoration.

\textbf{Real-world Images. }We further test the recovery performance of our incremental model on real-world images. As shown in Figure \ref{fig::real-visual}, our model still achieves better performances. Other methods either exhibit poor stability, characterized by significant forgetting of previous knowledge, or lack flexibility, resulting in insufficient learning of new knowledge. The visual results of AFC\cite{AFC2022} and our method are similar, but our method outperforms AFC~\cite{AFC2022} in the dehazing task. 
We also present some real-world scenes with combined degradations, such as rain and snow, all of the three, where our model outperforms other methods. However, there still contains some issues. In scenes with rain, haze, and snow, our model removes rain and snow, but some haze still remains.

\begin{table}[]
	\scriptsize
	\centering
    \caption{Comparisons among different distillation methods on top of the baseline method $\mathcal{L}_{SW}$ + $\mathcal{L}_{KD}$.}
	\setlength\tabcolsep{3.5pt}
	\scalebox{0.96}[1]{
		\begin{tabular}{ccccccc}
			\toprule
			\multirow{2}{*}{Methods}
			
			& \multicolumn{2}{c}{Average}
			& \multicolumn{2}{c}{Task1-OTS}
			& \multicolumn{2}{c}{Task2-Rain100H}
			\\
			\cmidrule{2-7} & PSNR & SSIM  & PSNR & SSIM  & PSNR & SSIM \\
			\midrule
                ER-$\mathcal{L}_{SW}$                      & 28.97    & 0.9322   & 29.12 & 0.9701  & 28.82 & 0.8942 \\
                $\mathcal{L}_{SW} + \mathcal{L}_{KD}$                & 29.21    & 0.9349   & 29.54 & 0.9724  & 28.89 & 0.8975  \\
                $\mathcal{L}_{SW} + \mathcal{L}_{KD}$ + Features      & 29.85    & 0.9313   & 31.42 & 0.9796  & 28.28 & 0.8831   \\
                $\mathcal{L}_{SW} + \mathcal{L}_{KD}$ + POD      & 29.28    & 0.9283   & 30.41 & 0.9763  & 28.16 & 0.8802   \\
                $\mathcal{L}_{SW} + \mathcal{L}_{KD}$ + AFC      & 29.12    & 0.9151   & 31.23 & 0.9788  & 27.00 & 0.8515   \\
                $\mathcal{L}_{SW} + \mathcal{L}_{KD}$ + SSRE      & 29.16    & 0.9340   & 29.44 & 0.9715  & 28.88 & 0.8965   \\
                $\mathcal{L}_{SW} + \mathcal{L}_{KD} + \mathcal{L}_{PKD}$    & \textbf{29.99}    & \textbf{0.9348}   & 31.35 & 0.9792  & 28.64 & 0.8903   \\
			\bottomrule
	\end{tabular}}
        \vspace{-6pt}
	\label{table: ablation}
\end{table}

\subsection{Ablation Study}

To analyze the effectiveness of the proposed method in detail, we conduct ablation experiments on the continual setting of dehazing$\rightarrow$deraining. We mainly analyze the effect of principal component projection, hyper-parameters in Eqn.~\eqref{overallLoss}, and the task order, on the continual learning model performances. 


\textbf{Principal component projection} is one important component in our knowledge replay based continual learning framework, which transfers knowledge from the intermediate network features of the previous task data, to mitigate \emph{catastrophic forgetting}. 
To analyze how this term helps to improve performance, we conduct the following analysis. 
As shown in Figure~\ref{fig::h}, (a) represents the input rainy image; (b) is the visualization of feature maps extracted by the original feature extractor $\mathcal{F}(\cdot)$, whose dimension is $W\times H \times 192$; (c) is the visualization of feature map further encoded by the principal component projection $\psi(\cdot)$ of $\mathcal{F}(\cdot)$, whose dimension is $W \times H \times 16$. We can clearly see that (c) remains the main information of the input feature while neglecting some redundant information. 
This principal component constraint helps to preserve principal knowledge of different tasks, and further achieve better plasticity-stability trade-off. 

Besides, we also conduct the following experiments on top of the baseline method $\mathcal{L}_{SW}$ + $\mathcal{L}_{KD}$, by further adding the following methods for comparisons: 1) Directly using the features for distillation ($\mathcal{L}_{SW}$ + $\mathcal{L}_{KD}$ + Features); 2) Using other methods with the baseline method, i.e., POD~\cite{Podnet2020}, AFC~\cite{AFC2022} and SSRE~\cite{SSRE2022}. Experiment results in Table~\ref{table: ablation} illustrate the efficiency of the proposed projection method.

\begin{table}[t]
    \scriptsize
    \centering
    \vspace{0pt}
    \caption{Experimental analysis of the effect of memory bank size on model performance. }
	\scalebox{0.88}[0.95]{
		\begin{tabular}{p{25pt}cccccccc}
			\toprule
			\multirow{2}{25pt}{Memory bank size} & \multicolumn{2}{c}{Average} & \multicolumn{2}{c}{Task1-OTS} & \multicolumn{2}{c}{Task2-Rain100H} & \multicolumn{2}{c}{Task3-Snow100K} \\
			\cmidrule{2-9} & PSNR & SSIM & PSNR & SSIM & PSNR & SSIM & PSNR & SSIM\\
			\midrule	
		  1000 & 30.72 & 0.9333 & 30.95 & 0.9773 & 28.63 & 0.8918 & 32.77 & 0.9346 \\   
            750 & 30.69 & 0.9339 & 30.72 & 0.9775 &	28.49 &	0.8886 & 32.83 & 0.9355   \\
           500 & 30.70 & 0.9335 & 31.03 & 0.9775 & 28.52 & 0.8900 & 32.54 & 0.9328 \\
            250 & 30.23 & 0.9314 & 29.29 & 0.9697 &	28.29 &	0.8863 & 33.12 & 0.9382 \\
            \bottomrule
	  \end{tabular}}
    \label{table: memory bank size}
\end{table}

\begin{table}[t]
    \scriptsize
    \centering
    \vspace{0pt}
    \caption{Experimental analysis of task order. }
	\scalebox{0.8}[0.95]{
		\begin{tabular}{ccccccccc}
			\toprule
			\multirow{2}{*}{Task order} & \multicolumn{2}{c}{Average} & \multicolumn{2}{c}{OTS} & \multicolumn{2}{c}{Rain100H} & \multicolumn{2}{c}{Snow100K} \\
			\cmidrule{2-9} & PSNR & SSIM & PSNR & SSIM & PSNR & SSIM & PSNR & SSIM\\
			\midrule	
		  haze$\rightarrow$rain$\rightarrow$snow & 30.70 & 0.9335 & 31.03 & 0.9775 & 28.52 & 0.8900 & 32.54 & 0.9328 \\   
            haze$\rightarrow$snow$\rightarrow$rain & 30.46 & 0.9331 & 30.18 & 0.9760 &	28.42 &	0.8875 & 32.79 & 0.9360   \\
           rain$\rightarrow$haze$\rightarrow$snow & 30.95 & 0.9404 & 30.11 & 0.9746 &	29.43 &	0.9067 & 33.29 & 0.9397 \\
            rain$\rightarrow$snow$\rightarrow$haze & 30.98 & 0.9393 & 30.88 & 0.9768 &	29.38 &	0.9060 & 32.68 & 0.9350 \\
            snow$\rightarrow$haze$\rightarrow$rain & 30.68 & 0.9361 & 30.19 & 0.9743 &	28.93 &	0.8957 & 32.93 & 0.9383 \\
            snow$\rightarrow$rain$\rightarrow$haze & 30.86 & 0.9372 & 30.69 & 0.9759 &	28.91 &	0.8967 & 32.99 & 0.9389 \\
            \bottomrule
	  \end{tabular}}
   \vspace{-10pt}
    \label{table: task order}
\end{table}

\textbf{Effect of the memory bank size} for knowledge replay. We further conduct experiments to explore the effect of memory bank size on the model performances. As shown in Table~\ref{table: memory bank size}, the quantity of stored old samples impacts the model's performances, with more stored samples achieving relatively better performances. When the memory bank size $M$=500, the model already exhibits results close to the upper bound, and further increases in the number of old samples do not significantly enhance the performance. We strike a balance between memory space and model performance, and finally choose $M$=500 in all the experiments.

\textbf{Effect of the task order} on the continual learning model has also been further investigated in Table~\ref{table: task order}. Specifically, we conduct experiments under all the possible six task order settings within these three task datasets. Experimental results show that the task order can slightly affect the model's performances, where the average PSNR/SSIM under different settings are relatively very close.


\begin{table}[t]
    \scriptsize
    \centering
    \vspace{0pt}
    \caption{Parameter sensitivity analysis with varying values of $\lambda$ under the setting of dehazing $\rightarrow$ deraining $\rightarrow$ desnowing. }
	\scalebox{0.88}[0.95]{
		\begin{tabular}{ccccccccc}
			\toprule
			\multirow{2}{*}{$\lambda$} & \multicolumn{2}{c}{Average} & \multicolumn{2}{c}{Task1-OTS} & \multicolumn{2}{c}{Task2-Rain100H} & \multicolumn{2}{c}{Task3-Snow100K} \\
			\cmidrule{2-9} & PSNR & SSIM & PSNR & SSIM & PSNR & SSIM & PSNR & SSIM\\
			\midrule	
		0.1 &30.51 	&0.9332 	&30.25 	&0.9745 	&28.43 	&0.8892 	    &32.86 	&0.9359   \\
            0.3 & \textbf{30.70} & \textbf{0.9335} & 31.03 & 0.9775 &	28.52 &	0.8900 & 32.54 & 0.9328   \\
            0.8 & 30.63 & 0.9324 & 31.22 & 0.9785 &	28.54 &	0.8898 & 31.98 & 0.9265 \\
            1.5 & 30.54 & 0.9309 & 31.29 & 0.9786 &	28.62 &	0.8905 & 31.71 & 0.9235 \\
            \bottomrule
	  \end{tabular}}
    \label{table: lambda}
\end{table}

\begin{table}[t]
    \scriptsize
    \centering
    \vspace{0pt}
    \caption{Parameter sensitivity analysis with varying values of $\alpha$ under the setting of dehazing $\rightarrow$ deraining $\rightarrow$ desnowing. }
	\scalebox{0.88}[0.95]{
		\begin{tabular}{ccccccccc}
			\toprule
			\multirow{2}{*}{$\alpha$} & \multicolumn{2}{c}{Average} & \multicolumn{2}{c}{Task1-OTS} & \multicolumn{2}{c}{Task2-Rain100H} & \multicolumn{2}{c}{Task3-Snow100K} \\
			\cmidrule{2-9} & PSNR & SSIM & PSNR & SSIM & PSNR & SSIM & PSNR & SSIM\\
			\midrule	
            0.1 & 30.49 & 0.9332 & 30.21 & 0.9751 &	28.49 &	0.8900 & 32.77 & 0.9344   \\
            0.5 & 30.68 & 0.9341 & 30.55 & 0.9759 &	28.39 &	0.8889 & 33.12 & 0.9377 \\
            1 & 30.70 & 0.9335 & 31.03 & 0.9775 &	28.52 &	0.8900 & 32.54 & 0.9328 \\
            1.5 & 30.64 & 0.9346 & 30.49 & 0.9769 &	28.55 &	0.8908 & 32.89 & 0.9361 \\
            \bottomrule
	  \end{tabular}}
    \label{table: alpha}
\end{table}

\begin{figure*}[ht] 
    \centering 
    \includegraphics[width=\textwidth]{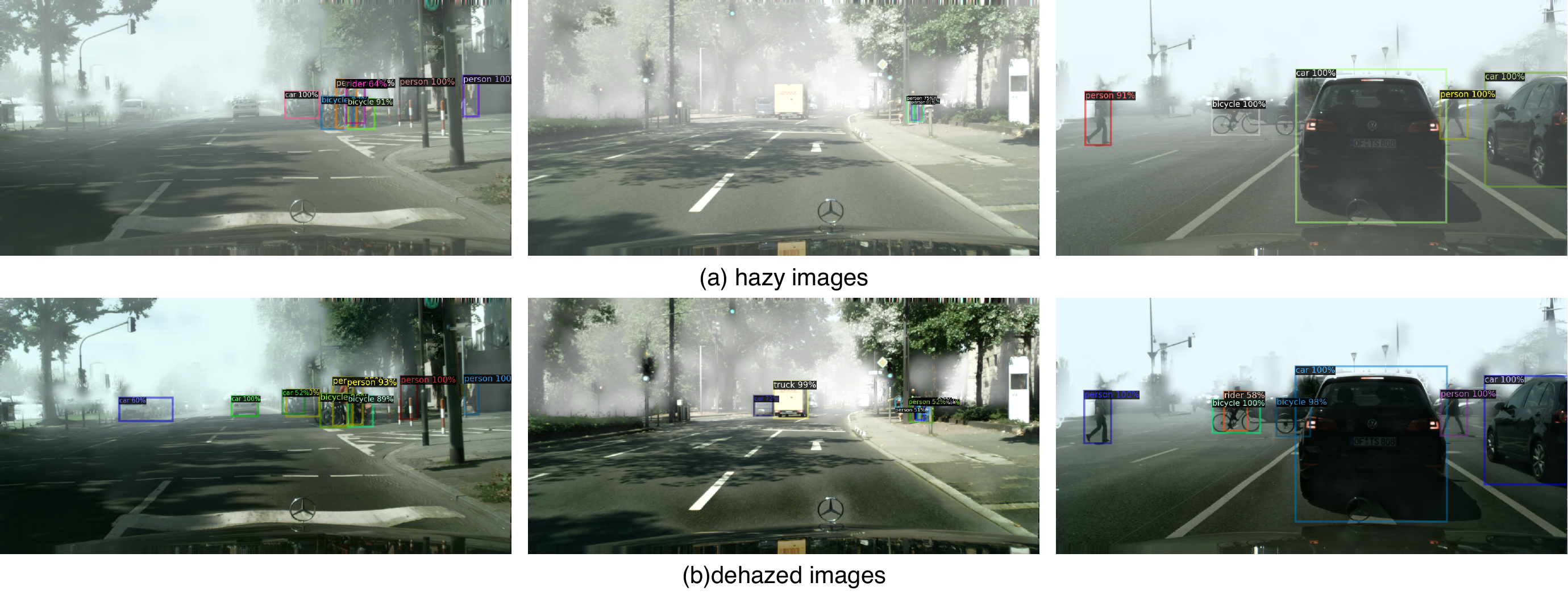}
    \captionsetup{justification=centering}
    \caption{ Visualization comparison of object detection on degraded images v.s. restored images by the proposed all-in-one adverse removal model. Specifically, a) is the detection results on the original hazy images; b) is the detection results on the restored dehazed images. } 
    \label{fig::object_detection}
    \vspace{-4pt}
\end{figure*}

\textbf{Parameter sensitivity analysis}. We have conducted detailed experiment analysis on the hyper-parameters of $\alpha$ and $\lambda$ in Eq.~\eqref{overallLoss}, which balance the three terms, $i.e.$, the reconstruction objective loss $\mathcal{L}_{SW}$, the knowledge replay distillation term $\mathcal{L}_{KD}$ and the principal knowledge distillation term $\mathcal{L}_{PKD}$.  Specifically, in Table~\ref{table: lambda}, it shows the parameter sensitivity analysis of $\lambda$ with varying values under the setting of dehazing$\rightarrow$deraining$\rightarrow$desnowing with $\alpha=1.0$, which controls the balance between the previous knowledge and the current task in the intermediate feature space. The experimental results indicate that the model tends to preserve old knowledge as $\lambda$ increases, and tends to learn new knowledge as $\lambda$ decreases. In Table~\ref{table: alpha}, it shows the parameter sensitivity analysis of $\alpha$ with varying values under the same setting with $\lambda=0.3$, which controls the balance between previous knowledge and current task in the output prediction. Overall, we can see that the model performance is not very sensitive to these two hyper-parameters, and the model yields the best performance when $\alpha=1.0$ and $\lambda=0.3$.


\begin{table}[ht]
    \centering
    \caption{Experiment comparison of our proposed method on object detection task.}
    \begin{tabular}{c|cc}
        \hline
        FoggyCityscape & Foggy images & Restored images \\
        \hline
        mAP & 26.36 & 29.30 \\
        \hline
    \end{tabular}
    \label{table:object-detection}
\end{table}

\subsection{Application on the Object Detection Task}
In order to illustrate the effect of our model pre-processed images on downstream computer vision tasks, we compare object detection results on degraded images v.s. images restored by the proposed method. Specifically, we adopt the commonly used Cityscape\cite{cityscapes} and FoggyCityscape\cite{foggycityscapes} datasets for performance evaluation. Each of these two datasets consists of 2,975 training and 500 validation images with 8 object categories: \emph{person, rider, car, truck, bus, train, motorcycle, and bicycle}. The images in the FoggyCityscape dataset\cite{foggycityscapes} are rendered from the Cityscape dataset by integrating fog and depth information. In the experiments, we adopt the Faster-RCNN{\cite{faster}} as the basic object detector, with pre-trained weights on the Cityscape dataset. The mean Average Precision (mAP) with IoU threshold of 0.5 is adopted as the evaluation metric. For comparison, we conduct the following two experiments: 1) we directly apply the pre-trained object detector on the test set of FoggyCityscape dataset; 2) we utilize our own pre-trained all-in-one adverse weather removal model to perform image restoration first on the test set of FoggyCityscape dataset, and then use the same object detector on these restored images. Experimental results are illustrated in Table~\ref{table:object-detection}, we can clearly see that image pre-processing by our all-in-one adverse weather removal model helps to improve the detection performance by a large margin of $2.94\%$mAP. Besides, we also select a subset of images for visualization, as depicted in Figure \ref{fig::object_detection}. It is evident from the visualizations that the restored images enhance visibility, leading to the detection of more targets.


\section{Conclusion}\label{sec:conclusion}
In this paper, we are the first to investigate continual learning for the all-in-one adverse weather removal task with multiple types of adverse weather degenerations in a setting closer to real-world scenarios. Specifically for the task, we develop a novel continual learning framework with a unified network structure and knowledge replay, where the knowledge replay techniques are tailored for the all-in-one adverse weather removal task by considering the characteristics of the image restoration task with multiple degenerations in CL. 
The proposed method performs competitively to existing dedicated or joint training image restoration methods, and we have built a benchmark for this task, which will be open-sourced to facilitate comparison for subsequent methods on this task.




%


{\small
\bibliographystyle{IEEEtran}
\bibliography{reference}
}

\end{document}